\documentclass[sigconf]{acmart}

\copyrightyear{2026}
\acmYear{2026}
\setcopyright{cc}
\setcctype{by}
\acmConference[HAI '26]{Proceedings of the 14th International Conference on Human-Agent Interaction}{November 16--19, 2026}{Osaka, Japan}
\acmBooktitle{Proceedings of the 14th International Conference on Human-Agent Interaction (HAI '26), November 16--19, 2026, Osaka, Japan}
\acmDOI{10.1145/3841580.3841599}
\acmISBN{979-8-4007-2575-3/2026/11}

\usepackage{balance}
\usepackage{booktabs}
\usepackage{xcolor}
\usepackage{tabularx}
\usepackage{colortbl}
\newcommand{\revision}[1]{\textcolor{black}{#1}}
\newcommand{\pid}[1]{\textsf{#1}}

\begin{document}

\title{From Tool to Agent: How Worker Needs Reorganize Across the Agentic Roles of Workplace AI}

\author{Christine P. Lee}
\orcid{0000-0003-0991-8072}
\affiliation{%
  \institution{Department of Computer Sciences University of Wisconsin--Madison}
  \country{Madison, Wisconsin, USA}
}
\email{cplee5@cs.wisc.edu}

\author{Min Kyung Lee}
\authornote{Equal senior contribution.}
\orcid{0000-0002-2696-6546}
\affiliation{%
  \institution{School of Information\\The University of Texas at Austin}
  \country{Austin, Texas, USA}
}
\email{minkyung.lee@austin.utexas.edu}

\author{Bilge Mutlu}
\authornotemark[1]
\orcid{0000-0002-9456-1495}
\affiliation{%
  \institution{Department of Computer Sciences University of Wisconsin--Madison}
  \country{Madison, Wisconsin, USA}
}
\email{bilge@cs.wisc.edu}

% ---------- Abstract ----------
\begin{abstract}
%As AI systems gain agency in workplaces, workers must increasingly absorb, explain, or execute decisions the AI makes on their behalf. We interviewed 16 workers in healthcare, finance, and management about their daily interactions with workplace AI. Across the three domains, AI occupied three distinct agentic roles: \emph{assistant}, supporting worker decision-making; \emph{arbiter}, making consequential decisions workers had to defend to customers; and \emph{manager}, directing work and displacing the social infrastructure through which work had been negotiated. Worker needs reorganized systematically with the configuration of decision authority and accountability between worker and AI: the same generic concern meant structurally different things in each role. We argue that the location of agency, not its presence, reorganizes worker experience: current deployments routinely place accountability on workers for outcomes whose authority has moved to the AI. We discuss design implications by configuration and a research agenda for treating agency configuration as a deliberate organizational design choice.

As AI systems gain agency in workplaces, workers increasingly interact with systems that do more than support tasks: they make decisions, allocate work, and shape how workers communicate with others. We interviewed 16 workers in healthcare, finance, and management about their daily interactions with workplace AI systems. Across these domains, AI occupied three agentic roles: \emph{assistant}, supporting worker decision-making; \emph{arbiter}, making consequential decisions that workers had to explain to downstream stakeholders; and \emph{manager}, directing work and displacing social structures through which work had been negotiated and learned. We find that worker needs depended on how decision authority and accountability were configured around the AI system. Explanation, control, accountability, and communication each took different meanings depending on whether AI supported, decided, or directed work. We contribute a role-based account of workplace human-agent interaction and identify accountability-authority asymmetry as a mechanism through which AI deployments can place responsibility on workers for outcomes whose authority has shifted to AI. We discuss implications for designing workplace AI around both the agent and its surrounding agency configuration.

\end{abstract}

% ---------- CCS Concepts and Keywords ----------




\maketitle

\section{Introduction}

Artificial intelligence (AI) systems are increasingly embedded in workplaces as more than tools for supporting individual tasks. They make lending decisions, allocate shifts, screen candidates, and recommend diagnoses \cite{eloundou2024gpts}. As AI systems take on work that human decision-makers previously performed, they increasingly operate as agents: systems that act toward goals with varying degrees of autonomy from human oversight~\cite{chan2023harms, feng2025levels, morris2024levels, mitchell2025fully}. This shift has prompted substantial attention in human-AI interaction (HAI) and adjacent fields, including taxonomies of agent autonomy~\cite{feng2025levels,kim2024taxonomy,morris2024levels}, mechanisms for delegating authority to agents~\cite{south2025authenticated}, and principles for maintaining meaningful human control~\cite{vandermeer2020meaningful, singh2022consent}.

Much of this work asks how agent autonomy should be designed, constrained, or delegated before deployment. However, workplace AI systems operate in settings that are rarely reducible to a single user-agent dyad. A system that approves or denies a loan, schedules a shift, or recommends a treatment is embedded in a broader structure of workers, customers, patients, managers, and institutions. The person interacting with the AI may not be the person who chose its autonomy level, benefits from its deployment, or bears formal responsibility for its consequences. Thus, the design question is not only how much autonomy an agent should have, but how that autonomy is positioned within an organizational workflow.

This distinction matters because many workplace AI deployments continue to encounter resistance, workarounds, and breakdowns. Clinical decision-support systems have struggled to gain adoption from clinicians~\cite{strickland2019watson, heaven2020google, henry2022teaming}; machine-learning systems in care have raised concerns about unintended consequences and overreliance~\cite{cabitza2017unintended,bucinca2021trust}; and algorithmic management systems have provoked sustained worker resistance in scheduling, logistics, and platform work~\cite{lee2015working,levy2022data,zhang2022algorithmic, milanez2025algorithmic}. Even well-intentioned fairness and explainability tools can miss the sociotechnical conditions under which AI systems are deployed~\cite{holstein2019improving,sambasivan2021everyone}. These problems are often discussed in terms of trust, transparency, fairness, or usability. We argue that they also reflect a structural issue: AI systems are placed into decision-making chains in ways that redistribute authority and accountability, often without corresponding support for the workers who must interpret, explain, or enact their outputs.

We examine this issue through interviews with 16 workers who interacted with workplace AI systems in healthcare, finance, and management. We sampled these domains because they provided contrasting relationships between AI systems and human decision-making. In healthcare, AI systems recommended or flagged information within clinical workflows. In finance, automated underwriting systems produced customer-facing approval decisions. In management, AI systems allocated, scheduled, or otherwise directed work. Our analysis examines how these surface differences corresponded to different configurations of decision authority and accountability around the AI system.

We identify three agentic roles that workplace AI occupied in our data: \emph{assistant}, \emph{arbiter}, and \emph{manager}. In the assistant role, AI supported workers who retained final decision authority and accountability. In the arbiter role, AI made consequential decisions that workers had to explain to customers. In the manager role, AI directed work or made recommendations that became the expected course of action. Across these roles, familiar worker concerns about explanation, control, accountability, and communication took different forms. For example, explanation supported expert verification in the assistant role, customer-facing accountability in the arbiter role, and contestability of system patterns in the manager role.

We use these findings to develop the concept of \emph{accountability-authority asymmetry}: a configuration in which workers remain accountable for outcomes whose authority has shifted partly or fully to the AI system. This asymmetry helps explain why the same high-level concern can require different design responses across workplace AI deployments. When workers decide and remain accountable, they need support for verification and expertise calibration. When AI decides but workers explain, they need explanations and alternatives that can be carried into downstream conversations. When AI directs work, workers need ways to inspect, contest, and discuss the system's patterns over time.

The contributions of this work are threefold. First, we provide an empirical typology of three agentic roles that workplace AI can occupy in deployed organizational workflows: assistant, arbiter, and manager. Second, we identify accountability-authority asymmetry as a mechanism linking the distribution of decision authority and accountability to role-specific worker needs. Third, we reframe workplace AI design from designing the agent in isolation to designing the agency configuration around the agent, with attention to where authority sits, where accountability lands, and what support workers need as a result.

\section{Related Work}

Our work connects three bodies of research: frameworks for designing agent autonomy, studies of worker preferences for AI involvement, and empirical studies of workers' experiences with deployed AI systems. We build on them by examining how worker needs vary across different configurations of AI authority and accountability.

\subsection{Designing Agent Autonomy}
A growing body of work treats autonomy as a property of AI agents that can be deliberately designed, calibrated, and constrained. This framing builds on human-factors models of automation, which identified types and levels of automation as design choices~\cite{parasuraman2000model}. \citet{feng2025levels} propose five levels of agent autonomy organized by the user's role: operator, collaborator, consultant, approver, and observer. \citet{morris2024levels} describe levels of general AI as combinations of capability and autonomy, while \citet{kim2024taxonomy} develop a taxonomy of robot autonomy for human-robot interaction. Across these frameworks, developers set agent autonomy first, and the human role follows.

Adjacent work examines how to manage the consequences of autonomous action. \citet{south2025authenticated} describe authenticated delegation, in which humans grant AI agents scoped permissions while maintaining accountability. Work on meaningful human control~\cite{vandermeer2020meaningful,mecacci2020meaningful} asks what is required for humans to remain morally responsible for decisions involving AI, building on earlier concerns about the responsibility gap created by autonomous learning systems~\cite{matthias2004responsibility}. \citet{chan2023harms} catalog harms specific to increasingly agentic systems, including accountability gaps that arise when authority and responsibility diverge. Similarly, \citet{sarkar2023enough} critiques ``collaboration'' as a dominant framing for human-AI interaction because it can confer agency on systems whose decisions humans absorb but do not co-produce. Recent empirical work has also begun to examine how users oversee agents in practice~\cite{dhanorkar2026human}.

This work provides important accounts of agent autonomy, governance, and oversight, but primarily centers the relationship between an agent and its immediate user. In workplace deployments, this framing is incomplete. The person interacting with the AI is not always the person who selected its autonomy level, benefits from its deployment, or bears institutional responsibility for its consequences. Our work extends autonomy frameworks by examining how agentic systems are experienced once placed inside organizational structures of authority and accountability.

\subsection{Worker Preferences for AI Involvement}
A complementary line of work asks what kinds of AI involvement workers prefer. \citet{shao2025future} introduce the Human Agency Scale, a five-level rubric of preferred human involvement from full automation to essential human control, and apply it across 844 tasks and 104 occupations. \citet{woodruff2024knowledge} report participatory workshops with 54 knowledge workers across seven industries, finding that workers often expect generative AI to perform menial work under human review while expressing concerns about deskilling and dehumanization; recent work on generative AI use in scientific organizations similarly shows that workplace AI adoption introduces both practical benefits and organizational risks~\cite{wagman2025generative}.

This literature establishes that workers have coherent preferences about AI involvement and that those preferences vary across tasks, workers, and contexts. However, preference studies primarily examine what workers want AI to do. Our study examines what happens when AI systems are already deployed and workers must operate within the authority and accountability structures those deployments create.

\subsection{Workers' Experience with Deployed AI}

Other works examine workers' experiences with deployed AI systems. Research on algorithmic management shows that workers often experience algorithmic systems as restrictive rather than enabling, and develop strategies of resistance, adaptation, or workaround~\cite{lynn2025regulating}. \citet{lee2015working} showed how data-driven management reshapes worker autonomy, while \citet{kellogg2020algorithms} theorized algorithms as a contested terrain of control. Recent work extends these analyses to broader workforce settings~\cite{jarrahi2023algorithmic,zhang2022algorithmic} and shows how warehouse workers engage in ``work games'' that subvert metrics while maintaining apparent compliance~\cite{cheon2025fulfillment}.

HCI and CSCW research has also examined deployed AI in specific high-stakes domains. Studies of clinical decision support show that workflow fit, expertise calibration, and uncertainty communication shape adoption, and that unintended consequences arise when these are not addressed~\cite{yang2019unremarkable,jacobs2021designing,cai2019hello,henry2022teaming,yildirim2024investigating,cabitza2017unintended}. Studies of child welfare and homeless services examine what frontline workers need from algorithmic decision support in high-stakes institutional settings~\cite{kawakami2022improving,kuo2023understanding}. Work with industry practitioners further shows that deployment conditions, not only model behavior, shape outcomes for affected people~\cite{holstein2019improving,sambasivan2021everyone,selbst2019fairness}. Recent workplace research also shows how AI adoption can reconfigure roles, relationships, responsibility, and power~\cite{cha2026values}.

These studies provide rich accounts of deployed AI, but they typically examine one domain or one configuration of AI involvement at a time. Our work links these literatures through a worker-side, cross-configuration account of workplace AI. We show that concerns such as explanation, control, accountability, and communication take different forms depending on how decision authority and accountability are distributed between AI systems and workers.

\section{Method}

\subsection{Theoretical Sampling and Domain Selection}
We recruited workers across three workplace domains: healthcare, finance, and management. We chose these domains theoretically rather than for breadth coverage, using them as contrasting cases of how AI systems are positioned in workplace decision-making~\cite{charmaz2014constructing}. Each domain involved AI systems positioned differently in the decision-making chain. In healthcare, clinical decision-support and computer-aided detection systems provided recommendations that clinicians evaluated before acting. In finance, automated underwriting systems generated loan approval or denial decisions that workers communicated to customers. In management, algorithmic scheduling, hiring, and resource-allocation systems directed worker activity or made recommendations that became the expected course of action. This contrast allowed us to examine how worker experiences varied with the position of AI agency in work.

We define a \emph{decision-support AI system} as a deployed system that uses machine learning, analytical algorithms, or rule-based reasoning to produce decisions or decision-relevant outputs at the point of work, and that workers interact with directly as part of their daily work. Four terms recur throughout the paper. \emph{Agency} describes the consequential role the system plays in work: whether its outputs suggest, decide, or direct action. \emph{Authority} describes who issues the consequential decision. \emph{Accountability} describes who must justify, explain, or absorb the consequences of that decision. \emph{Agentic role} names the configuration of these properties the AI occupies in a particular deployment; our three roles---assistant, arbiter, and manager---emerged from the configurations we observed. All systems in our study acted with agency in this worker-experience sense, regardless of how autonomous they were in a designer's technical sense. We use \emph{manager} structurally rather than literally: in this role, the AI output becomes the default directive around which worker behavior is organized, whether through formal oversight, workplace norms, or peer expectations.

\subsection{Participants and Recruitment}

Participants were recruited through university mailing lists and flyers posted in workplaces in the United States. A pre-screening survey confirmed eligibility based on domain, regular interaction with a decision-support AI system, and at least three months of system use. We recruited 16 workers: 6 in healthcare, 5 in finance, and 5 in management. Participants were aged 20--47 ($M=31.1$, $SD=8.2$); 9 identified as male and 7 as female. Table~\ref{tab:participants} summarizes the sample. Throughout the paper we use prefixes H, F, and M to denote healthcare, finance, and management participants respectively. These interviews constituted the first phase of a two-part session with each participant; the second phase involved a co-design activity reported in prior work \cite{lee2024ai}. The analyses and findings in this paper draw only on the interview data.

The study received ethical approval from the \revision{University of Wisconsin--Madison} institutional review board. Participants provided written informed consent before the interview and received \$50 USD upon completion of the study. 

\begin{table}[t]
\centering
\small
\caption{Participant demographics and AI system used. CAD = computer-aided detection; CDSS = clinical decision-support software; AUS = automated underwriting system; ASS = automated scheduling system; \revision{DSS = decision-support software.}}
\label{tab:participants}
{\renewcommand{\arraystretch}{1.15}%
\begin{tabularx}{\columnwidth}{@{} l c c >{\raggedright\arraybackslash}X l @{}}
\toprule
\textbf{ID} & \textbf{Age} & \textbf{Gender} & \textbf{Role} & \textbf{System} \\
\midrule
\rowcolor{gray!10} \pid{H1} & 34 & F & Radiologist & CAD \\
\rowcolor{gray!10} \pid{H2} & 26 & F & Acute Care Physician & CDSS \\
\rowcolor{gray!10} \pid{H3} & 26 & F & ER Physician & CDSS \\
\rowcolor{gray!10} \pid{H4} & 30 & F & Primary Care Physician & CDSS \\
\rowcolor{gray!10} \pid{H5} & 27 & M & Family Medicine Physician & CDSS \\
\rowcolor{gray!10} \pid{H6} & 46 & M & Radiologist & CAD \\
\pid{F1} & 25 & M & Banker & AUS \\
\pid{F2} & 35 & M & Loan Officer & AUS \\
\pid{F3} & 47 & M & Banker, Branch Manager & AUS \\
\pid{F4} & 31 & M & Banker, Branch Manager & AUS \\
\pid{F5} & 21 & F & Banker & AUS \\
\rowcolor{gray!10} \pid{M1} & 20 & F & Chain Crew Member & ASS \\
\rowcolor{gray!10} \pid{M2} & 23 & M & Chain Crew Member & ASS \\
\rowcolor{gray!10} \pid{M3} & 30 & M & HR Recruiting Director & DSS \\
\rowcolor{gray!10} \pid{M4} & 41 & F & Marketing Manager & DSS \\
\rowcolor{gray!10} \pid{M5} & 37 & M & Supply Chain Analyst & ASS \\
\bottomrule
\end{tabularx}}
\end{table}

\subsection{Interview Procedure}

Each interview lasted approximately 1.5 hours and took place in person at the participant's workplace. Interviews had three parts. First, participants described their day-to-day interactions with the AI system, including how it entered their workflow, what tasks it supported, and how decision-making was distributed between worker and system. Second, participants demonstrated their interaction with the system, walking through typical inputs, outputs, and follow-up actions. Third, participants reflected on benefits, drawbacks, challenges, workarounds, desired changes, information needs, and expectations for the system's future role. The pre-screening questionnaire and interview guide are available in supplementary materials.\footnote{The supplementary materials can be found at \url{https://osf.io/vr9bj/overview?view_only=32dd8ee115f94ac6b78959b573c0617c}}

\subsection{Analysis}

We analyzed the interviews using a grounded-theory-informed qualitative approach~\cite{glaser2017discovery,strauss1998basics,charmaz2014constructing} that combined inductive coding with cross-configuration comparison. Our goal was not to generate a general theory of workplace AI adoption, but to develop an analytic account of how workers experience AI systems positioned differently in organizational decision-making.

All interviews were audio-recorded, transcribed using automated transcription software, and manually checked for accuracy. The first author, who conducted the interviews, read all transcripts and wrote analytic memos documenting each participant's workflow, the AI system's role, and moments where workers described breakdowns, workarounds, or desired support.

We then conducted open coding to identify recurring concepts in workers' experiences with AI systems. Open codes captured worker actions, interaction breakdowns, perceived benefits and harms, explanation needs, uncertainty, control, communication, resistance, workarounds, and responsibility. This process produced 102 open codes with written definitions. For example, excerpts about unclear outputs, inability to explain system decisions, or trial-and-error attempts to change outcomes were coded as explanation needs and system opacity; excerpts about inability to influence, override, or correct outputs were coded as worker control.

To assess the clarity and consistency of open-code definitions, a second coder independently coded a subset of the data selected to span the three domains and major code families, including explanation needs, uncertainty, control, communication, resistance, and perceived responsibility \cite{mcdonald2019reliability}. The second coder was trained through multiple sessions of code definitions and example excerpts before independent coding. Agreement was high (Cohen's $\kappa=.86$). Disagreements were used to refine code definitions, clarify category boundaries, and identify codes that were too broad or overlapping. The second coder supported consistency in open-code application; the higher-level role typology was developed through comparative interpretation, memoing, and full team discussion.

%After open coding, we conducted axial coding to group codes into higher-level categories describing worker needs and frictions, including verification support, uncertainty communication, expertise calibration, downstream explanation, contextual override, fairness transparency, contestability, displaced communication, and workarounds. We used analytic memoing and diagramming to examine how these categories related to the AI system's position in the decision-making chain.

We then conducted axial coding to group codes into higher-level categories of worker needs and frictions, such as verification, downstream explanation, contestability, and displaced communication.

Finally, we conducted a cross-configuration comparison across the three domains. Rather than treating healthcare, finance, and management as separate work domains alone, we compared them as contrasting configurations of AI agency. We asked who issued the consequential decision, who could influence or contest it, who had to explain it to others, and who absorbed the consequences when it failed or created friction. This comparison led to the three agentic roles presented in the findings: assistant, where workers retained decision authority; arbiter, where AI made consequential decisions workers had to explain; and manager, where AI directed worker behavior or made recommendations the default around which work was organized.

Through this comparison, we identified \textit{accountability-authority asymmetry}: the extent to which workers remained accountable for outcomes whose decision authority had shifted to the AI system. By the final interviews within each domain, we were no longer identifying new role-specific need categories; \revision{we assessed this pattern through interview-level memos noting whether each interview added new codes or need categories.} We therefore treat the sample as sufficient for this role-based analytic account, while not claiming exhaustive saturation across workplace AI systems.
\section{Findings: Three Agentic Roles}
\label{sec:findings}

Worker needs in our data varied with how decision authority and accountability were configured around the AI system. Across the three domains, participants raised familiar concerns about explanation, control, accountability, and communication. However, these concerns took different forms depending on whether the AI supported a worker's decision, made a decision the worker had to explain, or directed work that workers were expected to follow. We organize the findings around three agentic roles that captured these configurations: assistant, arbiter, and manager. Together these roles show how the position of AI in the decision-making chain shaped the kinds of support workers needed.

\subsection{Three Configurations of Authority and Accountability}
\label{sec:roles-overview}

Across the three domains, AI systems differed in where they sat in the decision-making chain. We characterize these differences as three agentic roles: assistant, arbiter, and manager.

In healthcare, AI systems occupied an \emph{assistant} role. Clinical decision-support and computer-aided detection systems suggested diagnoses, flagged anomalies, or surfaced relevant information, while clinicians retained final decision authority and accountability. H6 described this relationship: ``The human is still the gold standard. We make the final call, and the AI is still in the background, trying to find its way in how to best assist us or fit in our workflow.''

In finance, AI systems occupied an \emph{arbiter} role. Automated underwriting systems generated approvals or denials that workers communicated to customers. In this configuration, authority for the consequential decision sat with the system, while workers remained responsible for explaining that decision to customers.

In management, AI systems occupied a \emph{manager} role. We use this term to describe a structural position in the workflow, not a single technical mechanism. We classified recommendation-producing systems as managers when following their output was institutionally expected and departing from it required justification; by contrast, in assistant cases, evaluating and overriding AI output was itself part of expected professional practice. Across the management cases, the AI output became the default directive around which worker action was organized. In all five management cases, the system was institutionally adopted as part of the organization's workflow, replacing or constraining prior manual practices. The mechanism through which this default shaped work differed: in scheduling and resource-allocation cases (M1, M2, M5), it was reinforced through assignments, oversight, and operational dependencies; in decision-support cases (M3, M4), it was reinforced through organizational expectations that workers use the system's recommendation as the starting point for professional decision-making. M1 described the formal version of this configuration: ``It's like a new manager. Now we just have someone that comes in once a week to check if we are properly following the system.'' M3 described the decision-support version: ``The AI recommendation becomes the starting point. Unless someone has a very strong reason to push back, we usually go with it. If you disagree, you have to explain yourself to the team.''

%We coded a recommendation-producing system as manager rather than assistant when following its output was institutionally expected and departing from it required justification; in the assistant cases, evaluating and overriding the output was itself the expected professional practice.

These roles are not intended to exhaust all possible workplace AI configurations. Rather, they provide an analytic structure for comparing how worker needs shifted in our data as AI systems took different positions in the decision-making chain. We next describe the needs that became salient within each role.

\subsection{Assistant: Expertise-Calibrated Support for Human Authority}
\label{sec:assistant}

In the assistant configuration, workers retained final decision authority and accountability. As a result, participants evaluated AI support in relation to their own expertise and decision-making practices. They did not describe explanation as a uniform need. Instead, they described different forms of support depending on whether the AI was helping them verify an expert judgment, reason through uncertainty, or learn how to make a decision. The central issue in this role was thus not only whether the AI explained its output, but whether its support was calibrated to the worker's expertise.

\paragraph{Verification support had to fit workers' reasoning practices.}
Healthcare workers described decision-making as oriented not only toward identifying the most likely answer, but also toward avoiding consequential errors. In this context, participants wanted AI systems to communicate the conditions under which their outputs might be unreliable, the features that shaped the recommendation, and the alternatives the system had considered. H6 explained: ``Ideally, I would love for the AI algorithm to say, ``My performance while looking for a bleed may be limited because of these factors, or your data doesn't match how I was trained.'' Information about quality factors like slice thickness and angulation, if there are caveats, would be helpful.'' Several participants also described wanting alternative outcomes or ``why not'' explanations rather than a single top recommendation. In these accounts, verification was not a matter of accepting the system's confidence score; it required integrating the system's reasoning with the worker's own clinical judgment.

\paragraph{The same explanation could create friction for experts and insufficiency for trainees.}
Participants also described different explanation needs depending on their level of expertise. Expert workers wanted compressed signals that could quickly support or challenge their existing judgment without adding unnecessary detail. Less experienced workers, by contrast, needed more expanded rationales that could support learning and help them understand how to reason through a case. The systems participants described did not distinguish between these needs. H6 noted: ``It's important to address `Who is using the AI?' Especially regarding expertise, what do they need to see? Right now, it's all the same.'' In the co-design phase of the same session, this participant prototyped two separate explanation designs, one for experts and one for trainees \cite{lee2024ai}. H2 described a related issue from her training time: the tools were too opaque to support early learning, so learning to use them came after she had developed independent judgment through other means.

These findings suggest that, in the assistant role, explanation needs were tied to the worker's position in an expertise trajectory. A single explanation interface could under-serve both experts and trainees: experts may experience detailed explanations as friction, while trainees may experience compressed outputs as insufficient guidance. Assistive AI thus needs to account not only for the task being performed, but also for the expertise relation it is entering.

\subsection{Arbiter: Workers as Explainers-of-Record}
\label{sec:arbiter}

In the arbiter configuration, the AI system produced consequential decisions, while workers remained responsible for communicating those decisions to customers. This created a different explanation problem than in the assistant role. Workers did not only need explanations to understand the system for themselves; they needed explanations that could be carried into customer-facing conversations. 
We characterize this position as becoming an \emph{explainer-of-record}: the worker became the human point through which an institutional AI decision was made answerable to affected people.

\paragraph{Workers answered for decisions they did not make.}
F5 described this position directly: ``The system is the one that approves or denies it, but I'm the person sitting in front of the customer. They're not asking the system why they were denied; they're asking me. So even though I didn't make the decision, I still have to explain it and help them figure out what to do next.'' Other finance participants described similar tensions. F3 explained: ``With the AI system, it can do what we used to do much faster and in larger quantities. But the system kind of leaves us in the dark, because we don't know what's going on, which we need to help customers.'' In this configuration, workers became the human face of an institutional decision, but had limited access to the reasoning needed to make that decision accountable to customers.

\paragraph{Explanation needs were downstream-facing, not only user-facing.}
The information workers wanted was shaped by the customer-facing work they had to perform after the AI produced a decision. F2 explained: ``Now part of my job is to educate, but with a system that just says ``no'' and ``refer with caution,'' it's not always clear. That makes it really hard for me to do that educational piece with clients.'' F2 also described having little to offer when apparently similar cases received different decisions: ``A friend referred a friend, very similar situations, but one got approved and the other denied. And I couldn't do much other than say ``Uh, I don't know man, I'm sorry.'''' Workers wanted information they could use to advise customers on possible next steps, including what factors shaped the denial and what changes might lead to a different outcome. As F5 put it, ``If it's not approved, how can I adjust it to fit that category? Which levers do I need to pull?''

Thus in the arbiter role, explanations had to be usable beyond the worker's own interaction with the system. They needed to support not only the worker's understanding of the system, but also the worker's ability to translate the decision into a meaningful customer-facing account.

\paragraph{Contextual override was needed to make accountability defensible.}
Workers also described cases where the system's decision appeared to miss important contextual information. F3 and F4, for example, described cases involving bankruptcies tied to medical-insurance events or natural disasters, where their professional experience suggested that the system's decision did not fully reflect the customer's circumstances. However, the system provided limited ways to register these exceptions, attach contextual information, or modify the recommendation. As a result, workers could be placed in the position of explaining a decision they could neither fully understand nor meaningfully repair.

This accountability-authority asymmetry was most visible in the arbiter configuration. Workers were not simply users of the AI system; they were intermediaries between institutional AI decisions and the people affected by them. Designs that treat the worker as the sole audience for explanation thus misidentify the explanation problem in this role: the explanation must support the worker's ability to make the AI decision answerable to someone else.

\subsection{Manager: From Coordination to Compliance}
\label{sec:manager}

In the manager configuration, AI systems structured worker action by making certain assignments, recommendations, or allocations the expected course of action. We use this role to describe a default-setting position in the workflow: the AI output shaped what workers were expected to do, even when the mechanism of enforcement differed across cases. Participants did not describe these systems only as decision aids. They described them as systems that changed how work was coordinated, how deviations were justified, and how workers interacted with managers and peers. Three needs were especially salient in this configuration: contestability of system patterns, transparency around allocation criteria, and preservation of communication channels that the AI had displaced.

\paragraph{Managerial AI made compliance the default and deviation costly.}
Across the management cases, workers described AI outputs as carrying a presumption of correctness. In formal scheduling and resource-allocation settings, this presumption was reinforced through managerial oversight. M2 explained: ``Someone comes in to check if we're following the system, and they kind of nag because it shows that we don't. Then we follow it for a while, but eventually, we just go back to what we want because we don't really like it.'' In decision-support settings, the system's recommendation shaped professional expectations within an institutionally adopted workflow. As M3 described, once the AI recommendation appeared, ``choosing something else feels like the decision that needs defending.'' These cases differed in how the AI output became authoritative, but workers described a similar asymmetry: following the system required little explanation, while deviating from it required justification.

\paragraph{Workers needed contestability at the pattern-level, not single outputs.}
Workers questioned not only individual assignments or recommendations, but recurring patterns in how the system organized work. M2, describing repeated station assignments at a multinational chain, asked: ``So we have three people come in at 4:30 and all of us get done at noon. So the question is, why is [worker A] always, always at the bar? \ldots{} I think it is related to who is faster or better. But is it actually?'' This concern was not about a single output in isolation. It was about whether the system was repeatedly distributing desirable and undesirable work in ways workers could not inspect or challenge. In this configuration, confidence metrics or one-off explanations were insufficient. Workers needed visibility into allocation criteria, data use, and patterns over time.

\paragraph{Managerial AI displaced communication channels that supported learning and repair.}
Participants also described losses in the social infrastructure surrounding work. Human managers and peers had previously served not only as sources of instruction, but as people workers could ask questions, negotiate exceptions, and learn from. When the AI absorbed parts of this coordinating role, workers had fewer reasons and fewer opportunities to engage in those conversations. M2 explained: ``With the AI system now, I miss the connections I had with managers or even other workers, because there is so much less interaction.'' In these accounts, the AI did not merely automate a managerial function; it altered the relational structure through which workers made sense of assignments, raised concerns, and learned workplace norms.

\paragraph{Different views and stakes made AI-directed work difficult to sustain.}
Managers and non-managerial workers often approached the system with different understandings of the AI-based work being coordinated. Managers valued the AI system for productivity, oversight, and coordination because they understood how individual deviations affected the broader workflow. Workers lower in the hierarchy, by contrast, often experienced the same AI system through their own assignments, where compliance could feel like accepting direction from a system they had little role in shaping. M5 described managerial frustration with workers who ignored the system: ``Some people just say, ``Well, I don't care, I'm gonna ignore it and do my own thing.'' But no, they need to know that when you don't follow the AI system's output, it breaks everything for everyone else.'' This account shows that resistance reflected more than individual dislike of the AI system. It also reflected different views and stakes across the hierarchy: managers understood non-compliance as disrupting interdependent work, while workers often saw the AI output mainly in relation to their own immediate tasks. In this configuration, breakdowns were difficult to resolve because the system asked workers to comply with coordination logic that they did not fully see, share, or have authority to shape.

These findings show that managerial AI did not simply automate coordination. It changed how coordination was made, justified, and negotiated across the workplace. Workers needed ways to inspect and contest system patterns, but they also needed organizational channels through which AI-directed work could be discussed and repaired. Without these channels, managerial AI risked turning coordination into compliance while weakening the social structures through which workers understood and shaped their work.

\subsection{Cross-Role Synthesis: Generic Needs, Role-Specific Meanings}
\label{sec:cross-role}
%Across the three roles, participants raised similar high-level concerns: explanation, control, accountability, and communication. However, these concerns referred to different forms of support depending on the agency configuration. In the assistant role, explanation supported workers' own verification and expertise calibration. In the arbiter role, explanation supported workers' ability to account for AI decisions to downstream stakeholders. In the manager role, explanation and transparency supported workers' ability to inspect and contest patterns of work allocation over time.

Across the three roles, common concerns---explanation, control, accountability, and communication---took different meanings depending on the agency configuration. In the assistant role, explanation supported verification and expertise calibration; in the arbiter role, downstream accountability; and in the manager role, pattern-level contestability. Table~\ref{tab:role-specific-needs} summarizes how these concerns were reorganized across the configurations in our data. Overall, worker needs were not generic reactions to AI opacity or automation, but reflected how authority and accountability were configured around the AI system.

%This synthesis helps explain why generic design responses may under-serve workers in particular workplace AI deployments. A more detailed explanation interface may support an assistant configuration, but it may not help an arbiter unless the explanation can be carried into a customer-facing conversation. Similarly, contestability in a manager configuration must operate at the level of allocation patterns and workplace norms, not only individual outputs. In our data, worker needs were therefore not generic reactions to AI opacity or automation; they were responses to how authority and accountability were configured around the AI system.

\begin{table*}[t]
\centering
\small
\caption{The same generic HAI concern means structurally different things across the three agentic roles. The role-specific meanings shape what would count as an adequate design response.}
\label{tab:role-specific-needs}
{\renewcommand{\arraystretch}{1.3}%
\begin{tabularx}{\textwidth}{@{} l >{\raggedright\arraybackslash}X >{\raggedright\arraybackslash}X >{\raggedright\arraybackslash}X @{}}
\toprule
\textbf{Generic concern} & \textbf{Assistant} & \textbf{Arbiter} & \textbf{Manager} \\
\midrule
\rowcolor{gray!10}
Explanation &
Helps workers verify against their own reasoning &
Helps workers explain to downstream stakeholders &
Helps workers contest patterns over time \\
Control &
Adjusts AI support to expertise &
Enables contextual override and case repair &
Challenges allocation rules and data use \\
\rowcolor{gray!10}
Accountability &
Worker decides and owns the outcome &
Worker answers for AI decisions to customers &
Worker complies with system-shaped outcomes \\
Communication &
Supports expert reasoning and learning &
Supports customer-facing translation &
Preserves workplace learning and coordination \\
\bottomrule
\end{tabularx}}
\end{table*}

\section{Discussion}
\label{sec:discussion}

Our findings show that worker needs in workplace AI deployments are shaped not only by what the AI system does, but by where it is positioned in the decision-making chain. Across the assistant, arbiter, and manager roles, familiar concerns about explanation, control, accountability, and communication took different forms depending on how decision authority and accountability were distributed between workers and AI systems. This suggests that workplace HAI should treat agency not only as a property of the agent, but as a configuration of authority, accountability, and worker involvement around the agent.

\subsection{Agency Configuration as a Unit of Analysis}

Existing autonomy frameworks characterize what an agent can do and what role a human user plays in relation to the agent~\cite{feng2025levels,morris2024levels,kim2024taxonomy}. Worker-preference studies map what level of AI involvement workers want across tasks~\cite{shao2025future,woodruff2024knowledge}. Deployment studies show how workers experience AI systems in specific domains~\cite{lee2015working,cheon2025fulfillment,kawakami2022improving}. Our findings thus identify \emph{agency configuration} as a unit of analysis: the arrangement of decision authority, accountability, and worker involvement around a deployed AI system.

This distinction matters because authority and accountability can diverge. In the assistant role, workers retained both authority and accountability; their needs centered on verification, uncertainty, and expertise calibration. In the arbiter role, authority shifted to the AI while workers remained accountable to customers; their needs centered on explanations and alternatives they could carry into downstream conversations. In the manager role, AI outputs structured expected worker behavior; workers needed ways to inspect, contest, and discuss system patterns over time. These differences suggest that autonomy level alone does not fully describe workers' experience of agentic AI. A worker may appear to be an approver, observer, or recipient of an AI output, but their practical needs depend on whether they must also justify, repair, or absorb the consequences of that output.

We describe this pattern as \emph{accountability-authority asymmetry}. Concurrent work on algorithmic management reports a related asymmetry among frontline supervisors, who absorb accountability for outcomes they can neither fully understand nor contest \cite{ozdemir2026caught}. Our findings complement this observation by showing how the relationship between authority and accountability varies across agentic configurations, including cases in which workers must explain AI decisions to customers. The concept helps explain why similar high-level concerns can require different forms of support across deployments. Explanation in the assistant role supports workers' own reasoning; explanation in the arbiter role supports customer-facing accountability; and transparency in the manager role supports pattern-level contestability. Thus, the design problem is not only whether explanation or control is provided, but whether it matches the worker's agency configuration.

\subsection{Designing for Role-Specific Worker Needs}

The three roles point to different design priorities. In the assistant configuration, existing work on calibrated trust, appropriate reliance, cognitive forcing, sociotechnical explanation, and elicitation of explanation needs offers useful foundations~\cite{lee2004trust,bansal2021whole,bucinca2021trust,ehsan2021expanding,jacobs2021designing,liao2020questioning,miller2019explanation, lee2024ai}. Our findings suggest that these supports should be calibrated to worker expertise. Expert workers may need compressed signals for fast verification, while trainees may need more detailed rationales that support learning. Systems that pair automated verification with user-adjustable constraints show one way to support checking without displacing the worker's own judgment \cite{lee2025veriplan, lee2026u}. This concern aligns with recent work on dignified human-AI interaction, which warns that AI systems can improve immediate performance while weakening expertise and agency over time \cite{ehsan2026future}.
Uncertainty cues should also communicate when and why the system may be unreliable, not only the system's confidence~\cite{cabitza2017unintended,henry2022teaming}. This shifts the design question from whether to explain AI outputs to how explanation depth, uncertainty, and workflow support should vary across expertise levels.

The arbiter configuration is less directly served by existing explainable AI (XAI) approaches. Contestable-AI work and social-explanation research often treat the worker as the audience for the system's account~\cite{alfrink2023contestable,kawakami2022improving,miller2019explanation}. Our findings reposition the worker as an intermediary whose audience is downstream, echoing recent work on AI in street-level bureaucracy and social-service contexts where frontline workers mediate institutional decisions for affected publics \cite{busch2025artificial, karusala2024understanding}. Designs for arbitral AI should therefore support relay-capable explanations: rationales, alternatives, and counterfactuals that workers can use in customer-facing conversations. In the co-design phase of our study, finance workers designed explanations whose intended audience was the customer rather than themselves, including counterfactuals showing which changes would produce a different outcome \cite{lee2024ai}. They should also support contextual override or case repair when workers identify circumstances the system did not account for, treating repair as a designed interaction rather than an informal workaround \cite{lee2024rex}. In this configuration, explanation is not only a resource for understanding the AI; it is a resource for making institutional decisions answerable to affected people.

The manager configuration requires design beyond the individual interface. Research on algorithmic management has documented how AI systems can intensify control, constrain autonomy, and generate worker resistance~\cite{lee2015working,kellogg2020algorithms,jarrahi2023algorithmic,cheon2025fulfillment}. Work on industry practice and sociotechnical fairness further shows that interface-level fixes alone cannot address deployment conditions~\cite{holstein2019improving,selbst2019fairness,sambasivan2021everyone}. Our findings suggest three design priorities: pattern-level transparency into allocation and data-use patterns~\cite{zhang2022algorithmic}, collective contestability for challenging recurring system behavior, and communication structures that preserve peer learning, mentoring, and negotiation. Recent work on AI agents that stand in for managers reports the same boundary from the other direction: workers accepted such agents as supplements to managerial communication but rejected them as substitutes for it \cite{hu2026your}. These priorities are increasingly relevant as algorithmic management expands beyond platform work into conventional workplaces \cite{milanez2025algorithmic}. These priorities treat managerial AI not only as an interface problem, but as an organizational design problem.

\subsection{Toward Deliberate Agency Configuration}

A practical implication of our findings is that agency configuration should be treated as a deliberate design and deployment decision. In the workplaces we studied, AI systems were often positioned for productivity, efficiency, or coordination, while the worker's resulting role emerged after deployment. This can create accountability-authority asymmetry by default: workers become responsible for interpreting, explaining, or enacting AI outputs without the authority or support needed to do so effectively.

Before deployment, designers and organizations should ask three questions. First, where in the decision-making chain is the AI being placed? Second, where will accountability for the AI's outputs sit relative to that placement? Third, what worker needs follow from that configuration? These questions can complement existing autonomy frameworks, worker-preference instruments, and meaningful human control approaches~\cite{feng2025levels,shao2025future,mecacci2020meaningful,vandermeer2020meaningful,south2025authenticated,matthias2004responsibility}. They also align with sociotechnical critiques that caution against treating AI design as a technical fix detached from deployment context~\cite{selbst2019fairness}. Because worker participation in deployment decisions is often structurally constrained~\cite{lee2015working,kellogg2020algorithms,cheon2025fulfillment,kawakami2022improving}, treating configuration as a design choice also requires asking who has the authority to make that choice.

%Our findings open three directions for future HAI research. First, the role typology should be tested in settings where domain and configuration vary independently, such as assistant-style AI in finance or manager-style AI in healthcare. Second, arbitral AI calls for multi-stakeholder work that includes the downstream audiences of AI decisions, such as customers, patients, or applicants. Third, managerial AI calls for longitudinal work on whether displaced social infrastructure can be preserved through deliberate organizational design. These directions shift attention from designing the agent alone to designing the configuration in which the agent acts.
\section{Limitations and Future Work}

Our study has several limitations. First, our data come from 16 participants across three workplace domains, selected as theoretically contrasting cases rather than for exhaustive coverage. The three agentic roles should therefore be treated as an analytic account rather than an exhaustive taxonomy. Moreover, role and domain are confounded in our sample, with each role corresponding primarily to one domain. We cannot empirically disentangle whether observed needs arose from the agency configuration, the domain, or their interaction. Future work should test the typology in settings where these vary independently, including arbiter-style systems beyond automated underwriting, and examine whether different mechanisms through which AI becomes authoritative produce distinct worker needs.

Second, our analysis reflects workers' firsthand experiences within U.S. workplaces. We did not interview organizational decision-makers, system designers, or downstream stakeholders such as customers and patients, and institutional differences in labor regulation, employment protections, and workplace norms may shape how authority and accountability are distributed. Multi-stakeholder, longitudinal, and cross-cultural studies are therefore needed to examine how agency configurations are established, contested, and experienced across organizational contexts.

\section{Conclusion}

As AI systems take on more consequential roles in workplaces, worker needs depend not only on what the system does, but on where it is positioned in the decision-making chain. Through interviews with 16 workers across healthcare, finance, and management, we identified three agentic roles of workplace AI: assistant, arbiter, and manager. Across these roles, familiar concerns about explanation, control, accountability, and communication took different forms depending on how decision authority and accountability were distributed between workers and AI systems. We introduced accountability-authority asymmetry to describe configurations in which workers remain responsible for outcomes whose authority has shifted to AI. These findings suggest that workplace HAI should treat agency configuration as a deliberate design concern. Designing responsible workplace AI requires designing not only the agent, but also its surrounding authority, accountability, and worker supports.

%We examined how 16 workers across healthcare, finance, and management experience workplace AI systems. The systems occupied three agentic roles: assistant, arbiter, and manager. Worker needs reorganized systematically across these roles in ways that tracked the distribution of decision authority and accountability between the worker and the AI. We argued that the location of agency, not its presence, is what reorganizes worker experience, and that current deployments routinely create asymmetries in which workers carry accountability for outcomes whose authority has been moved to the AI. Addressing the workplace AI moment seriously will require designing not only the agent but the configuration in which it sits.

\begin{acks}
This research was supported by the National Science Foundation through awards IIS-1925043, IIS-1939606, DGE-2125858, CCF-2217721; Good Systems, a UT Austin Grand Challenge for developing responsible AI technologies;\footnote{https://goodsystems.utexas.edu} and the Swedish Research Council. Any opinions, findings, conclusions, or recommendations expressed in this material are those of the authors and do not necessarily reflect the views of the National Science Foundation.
\end{acks}

\balance
\bibliographystyle{ACM-Reference-Format}
\bibliography{references}

\end{document}